\documentclass[11pt,a4paper]{article}
\usepackage[hyperref]{acl2020}
\usepackage{times}
\usepackage{latexsym}

\usepackage{microtype}

\usepackage{amsmath}
\usepackage{amssymb}
\usepackage{graphicx}
\usepackage{hyperref}

\aclfinalcopy 

\title{TripPy: A Triple Copy Strategy for Value Independent\\Neural Dialog State Tracking}

\author{Michael Heck, Carel van Niekerk, Nurul Lubis,\\\textbf{Christian Geishauser, Hsien-Chin Lin, Marco Moresi, Milica Ga\v{s}i\'{c}} \\
  Heinrich Heine University D{\"u}sseldorf, Germany \\
  \texttt{\{heckmi,niekerk,lubis,geishaus,linh,moresi,gasic\}@hhu.de}}

\date{}

\begin{document}
\maketitle
\begin{abstract}
Task-oriented dialog systems rely on dialog state tracking (DST) to monitor the user's goal during the course of an interaction. Multi-domain and open-vocabulary settings complicate the task considerably and demand scalable solutions. In this paper we present a new approach to DST which makes use of various copy mechanisms to fill slots with values. Our model has no need to maintain a list of candidate values. Instead, all values are extracted from the dialog context on-the-fly. A slot is filled by one of three copy mechanisms: (1) Span prediction may extract values directly from the user input; (2) a value may be copied from a system inform memory that keeps track of the system's inform operations; (3) a value may be copied over from a different slot that is already contained in the dialog state to resolve coreferences within and across domains. Our approach combines the advantages of span-based slot filling methods with memory methods to avoid the use of value picklists altogether. We argue that our strategy simplifies the DST task while at the same time achieving state of the art performance on various popular evaluation sets including MultiWOZ 2.1, where we achieve a joint goal accuracy beyond 55\%.
\end{abstract}

\section{Introduction}
\label{sec:introduction}

The increasing popularity of natural language human-computer interaction urges the development of robust and scalable task-oriented dialog systems. In order to fulfill a user goal, a dialogue system must be capable of extracting meaning and intent from the user input, and be able to keep and update this information over the continuation of the dialog~\cite{young2010hidden}. This task is called dialog state tracking (DST). Because the next dialog system action depends on the current state of the conversation, accurate dialog state tracking (DST) is absolutely vital.

DST is tasked to extract from the user input information on different concepts that are necessary to complete the task at hand. For example, in order to recommend a restaurant to a user, the system needs to know their preferences in terms of price, location, etc. These concepts are encapsulated in an ontology, where dialogue domain (e.g., restaurant or hotel), slot (e.g., price range or location), and value (e.g. cheap or expensive) are defined. Solving this information extraction task is prerequisite for forming a belief over the dialog state.

\begin{figure}[t]
	\centering
	\includegraphics[page=1, trim=2.7cm 0.5cm 2.5cm 1.3cm, clip=true, width=1.00\linewidth,]{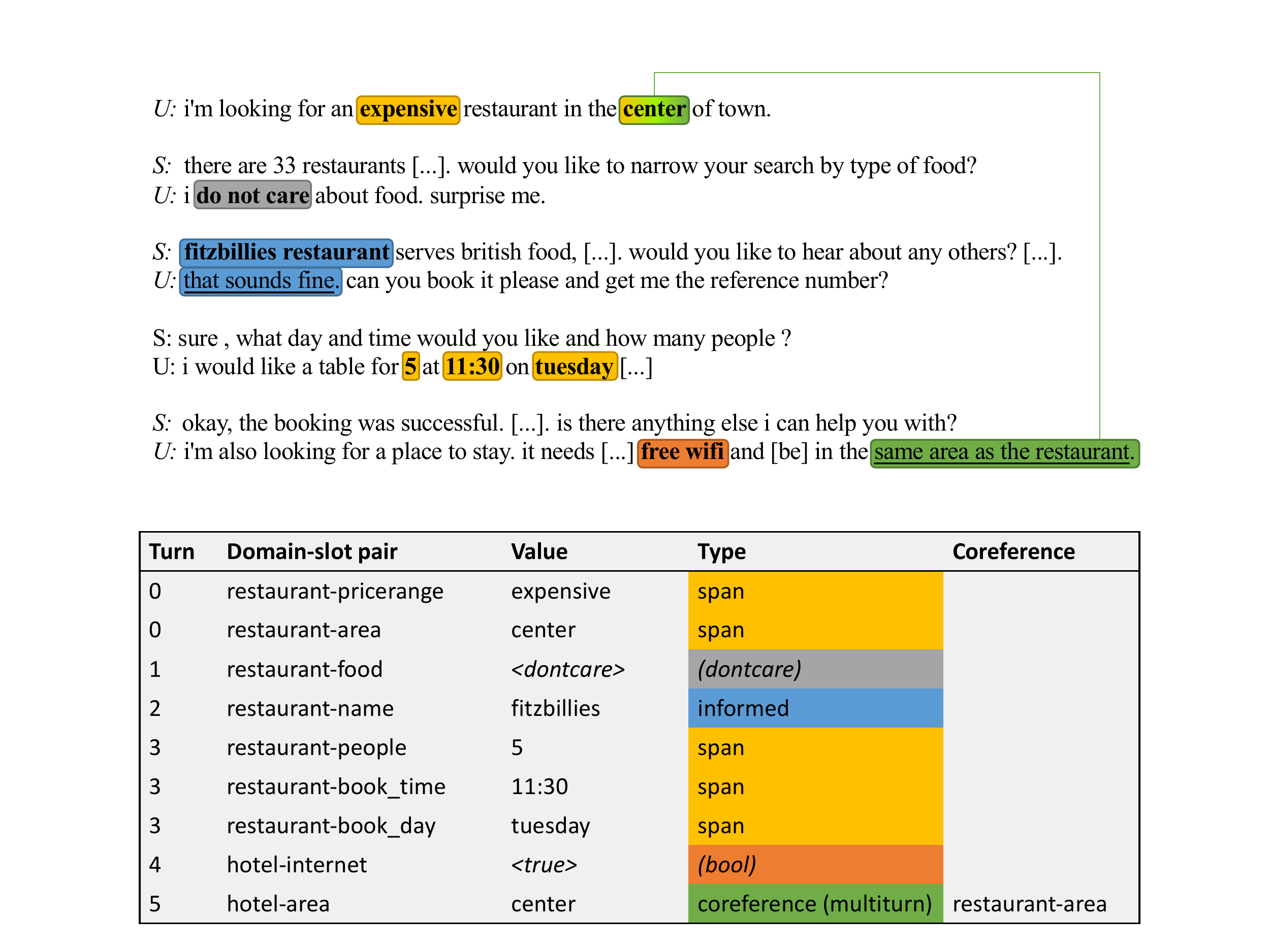}
    \caption{Example dialog in MultiWOZ.}	\label{fig:diag}
\end{figure}

Traditional approaches to DST operate on a fixed ontology and perform prediction over a pre-defined set of slot-value pairs~\cite{mrkvsic2016neural,liu2017end,zhong2018global}. Such approaches perform very well on datasets which are defined over fairly small ontologies. Apply these methods to more complex datasets however reveals various limitations~\cite{ren2018towards,nouri2018toward}. First, it is often difficult to obtain a complete ontology for a task. Second, slot-value pairs that were outside the ontology or the training data are impossible to capture during test time. Third, such methods at best scale linearly with the size of the ontology. Most importantly, the idea of fixed ontologies is not sustainable, as in real world applications they are subject to constant change.

Human-computer interactions often need to be defined over multiple domains at the same time, ideally with unrestricted vocabulary. Recent approaches to multi-domain and open-vocabulary DST extract values from the dialog context directly by predicting value spans in the input~\cite{gao2019dialog,chao2019bert,kim2019efficient,zhang2019find}. Span prediction is a demonstrably potent method to detect relevant information in utterances, but its major drawback is that it only suits \emph{extractive} values that are explicitly expressed as a sequence of tokens. This is the reason why span-based methods benefit from the support of a picklist, i.e., a list of value candidates from which a system can choose. Still, these methods fall short when handling nuanced and subtle phenonema that often occur in natural conversations, such as coreference and value sharing ("I'd like a hotel in the same area as the restaurant."), and implicit choice ("Any of those is ok."). 

In this work, we propose a new approach to value independent multi-domain DST:

\begin{enumerate}
    \item In addition to extracting values directly from the user utterance via span prediction and copy, our model creates and maintains two memories on-the-fly, one for system inform slots, and one for the previously seen slots.
    \item The \emph{system inform memory} solves the implicit choice issue by allowing to copy from concepts mentioned by the system, e.g., values that are offered and recommended.
    \item The \emph{DS memory} allows the use of values already existing in the dialogue state to infer new values, which solves the coreference and value sharing problems.
\end{enumerate}
We call this approach \textbf{TripPy}, \textbf{Trip}le co\textbf{py} strategy DST.\footnote{Our code is available at \url{https://gitlab.cs.uni-duesseldorf.de/general/dsml/trippy-public}.} Our experiments results show that our model is able to handle out-of-vocabulary and rare values very well during test time, demonstrating good generalization. In a detailed analysis we take a closer look at each of the model's components to study their particular roles.

\section{Related Work}

Dialog state tracking has been of broad interest to the dialog research community, which is reflected by the existence of a series of DST challenges~\cite{henderson2014second,rastogi2019towards}. These challenges consistently pushed the boundaries of DST performance. Current state-of-the-art has to prove to work on long, diverse conversations in multiple domains with a high slot count and principally unrestricted vocabulary~\cite{eric2019multiwoz}. Dialogs of such complex nature are tough for traditional approaches that rely on the availability of a candidate list due to scalability and generalization issues~\cite{mrkvsic2016neural,liu2017end,ramadan2018large,rastogi2017scalable}.

Span-based approaches recently alleviated both problems to some extent. Here, slot values are extracted from the input directly by predicting start and end positions in the course of the dialog. For instance,~\citet{xu2018end} utilizes an attention-based recurrent network with a pointer mechanism to extract values from the context. This extractive approach has its limitations, since many expressible values are not found verbatim in the input, but rather mentioned implicitly, or expressed by a variety of rephrasings.

With the assistance of contextual models such as BERT~\cite{devlin2018bert}, issues arising from expressional variations can be mitigated. Recent work has demonstrated that encoding the dialog context with contextual representations supports span prediction to generalize over rephrasings. SUMBT~\cite{lee2019sumbt} utilizes BERT to encode slot IDs and candidate values and learns slot-value relationships appearing in dialogs via an attention mechanism. Dialog context is encoded with recurrence. BERT-DST~\cite{chao2019bert} employs contextual representations to encode each dialog turn and feeds them into classification heads for value prediction. The dialog history, however, is not considered for slot filling. In~\citet{gao2019dialog}, DST is rendered as a reading comprehension task that is approached with a BERT-based dialog context encoder. A slot carryover prediction model determines whether previously detected values should be kept in the DS for the current turn.

An alternative to span prediction is value generation. TRADE~\cite{wu2019transferable} and MA-DST~\cite{kumar2020ma} generate a DS from the input using a copy mechanism to combine the distributions over a pre-defined vocabulary and the vocabulary of current context. SOM-DST~\cite{kim2019efficient} applies a similar mechanism for value generation, but takes the previous dialog turn as well as the previous DS as input to BERT to predict the current DS. A state operation predictor determines whether a slot actually needs to be updated or not. The downside of generative models is that they tend to produce invalid values, for instance by word repetitions or omissions.

Recently, a hybrid approach called DS-DST has been proposed that makes use of both span-based and picklist-based prediction for slot-filling~\cite{zhang2019find}. In contrast to generative approaches, picklist-based and span-based methods use existing word sequences to fill slots. DS-DST somewhat alleviates the limitations of span prediction by filling a subset of slots with a picklist method instead.

Recent works seemed to reveal a trade-off between the level of value independence in a model and the DST performance. \citet{chao2019bert} and~\citet{gao2019dialog} solely rely on span-prediction, but their performance lacks behind methods that at least partially rely on a pre-defined list of candidate values. This has impressionably been demonstrated by~\citet{zhang2019find}. Their model could not compete when relying on span-prediction entirely. In contrast, when relying solely on their picklist slot-filling method, they achieved the to-date best performance on MultiWOZ 2.1. The proposed dual-strategy approach lies favorably between these two extremes.

To the best of our knowledge, none of the recent approaches to complex DST tasks such as MultiWOZ~\cite{budzianowski2018multiwoz,eric2019multiwoz} are value independent in the strict sense. What's more, they tremendously benefit from the use of a value candidate list. Our work tackles this limitation by introducing a triple copy strategy that relies on span-prediction as well as memory mechanisms. In contrast to other hybrid approaches such as~\citet{zhang2019find}, our memory mechanisms create candidate lists of values on-the-fly with the dialog context as only source of information, thus avoiding the use of pre-defined picklists. 
We let the model decide which strategy to choose for each slot at each turn. Our approach differs from~\citet{chao2019bert} and~\citet{kim2019efficient} in that we consider the dialog history as context in addition to the current turn. We also differ from approaches like~\citet{lee2019sumbt} since we do not employ recurrence. Like~\citet{kim2019efficient}, we use auxiliary inputs at each turn, but we do so as a late feature fusion strategy. With our slot-value copy mechanism to resolve coreferring value phrases, we employ a method which is reminiscent of~\citet{gao2019dialog}'s slot carryover, but with the sharp distinction that we copy values between different slots, facilitating value sharing within and across domains.

\begin{figure*}[t]
	\centering
	\includegraphics[page=1, trim=0.0cm 1.5cm 0.5cm 2cm, clip=true, width=1.00\linewidth,]{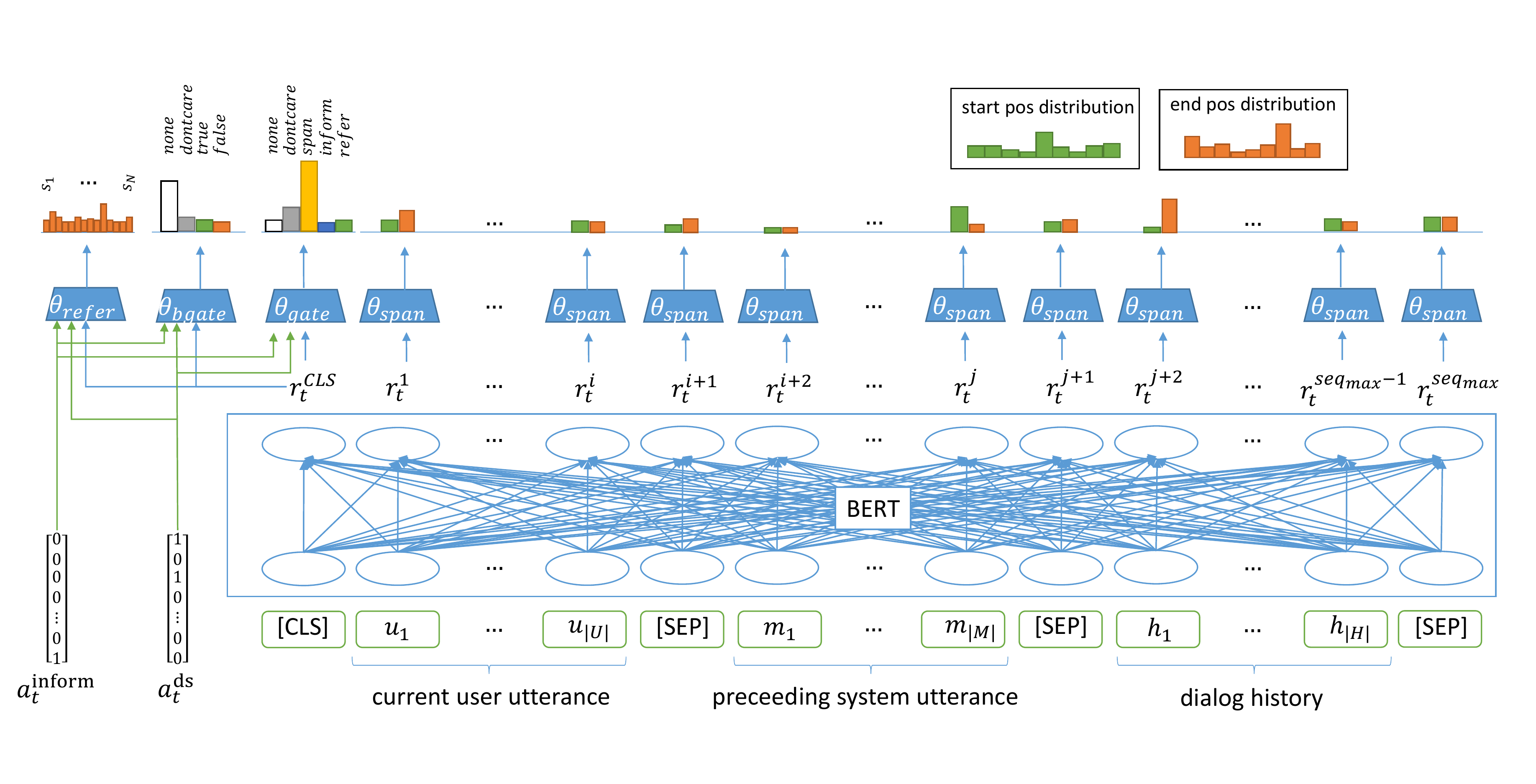}
	\caption{Architecture of our proposed model. TripPy takes the turn and dialog history as input and outputs a DS.}
	\label{fig:model}
\end{figure*}

\section{TripPy: Triple Copy Strategy for DST}

Our model expects the following input format to perform dialog state tracking. Let $X = \{(U_1, M_1), \dots, (U_T, M_T)\}$ be the sequence of turns that comprise a dialog of length $T$. $U_t$ is the user utterance at turn $t$, $M_t$ is the system utterance that preceeds the user utterance. The task of the model is (1) to determine for every turn whether any of the $N$ domain-slot pairs in $S = \{S_1, \dots, S_N\}$ is present, (2) to predict the values for each $S_n$ and (3) to track the dialog state $DS_t$ over the course of the dialog, i.e., for $t \in [1, T]$.

We employ a triple-copy strategy to fill the slots. The intuition is that values are either explicitly expressed by the user, that they are expressed by the system and referred to by the user via confirmation or rejection, or that they have been expressed earlier in the dialog as assignment to another domain-slot pair (coreference). Each of these cases is handled by one of three copy mechanisms. It becomes apparent that slots can not be filled by exclusively resorting to one particular copy method. Therefore, we employ slot gates that determine at each turn which method to use to fill the respective slot. 

Figure~\ref{fig:model} depicts our model. We encode the dialog context with a BERT front-end and feed-forward the resulting contextual representations to various classification heads to solve the sub-tasks for DST. The aggregate sequence representation is the input to the slot gates. The sequence of token representations is the input to the span predictors.

\subsection{Context Encoder}

We use BERT~\cite{devlin2018bert} as front-end to encode at each turn $t$ the dialog context as
\begin{equation}
\begin{split}
    R_t = \mathrm{BERT}(&\mathrm{[CLS]} \oplus U_t \oplus \mathrm{[SEP]} \oplus M_t \oplus \\
                        &\mathrm{[SEP]} \oplus H_{t} \oplus \mathrm{[SEP]}),
\end{split}
\end{equation}
where $H_t = {(U_{t-1}, M_{t-1}), \dots, (U_1, M_1)}$ is the history of the dialog up to and excluding turn $t$. The special token [CLS] preceeds every input sequence to BERT, and [SEP] separates portions of the input sequence. It is then
$R_t = [r_t^{\mathrm{CLS}}, r_t^1, \dots, r_t^{\mathrm{seq_{max}}}],$
where $r_t^{\mathrm{CLS}}$ is a representation of the entire turn including the dialog context $H_t$. The vectors $r_t^1$ to $r_t^{\mathrm{seq_{max}}}$ are contextual representations for the sequence of input tokens (including special tokens). Both types of representations are used for the following classification tasks.

\subsection{Slot Gates}

Our model is equipped with a slot gate for each domain-slot pair. This ensures greatest flexibility for multi-domain DST, as there is no restriction as to how many domains might be present in a single turn.
At each turn $t$, slot gates assign each slot $S_n$ to one of the classes in $C = \{\mathit{none, dontcare, span, inform, refer}\}$. The first two labels express special cases. $none$ denotes that the slot does not take a value in this turn and $\mathit{dontcare}$ states that any value is acceptable for this slot. The remaining three labels each denote one of the model's copy mechanisms. $span$ indicates that a value is present in $U_t$ that can be extracted via span prediction. $\mathit{inform}$ indicates that the user refers to a value that has been uttered by the system in $M_t$. Lastly, $\mathit{refer}$ indicates that the user refers to a value that is already present in $DS_t$.

The input to the slot gates is $r_t^{\mathrm{CLS}}$, and the probability distribution over classes $C$ for domain-slot pair $S_n$ at turn $t$ is $p^\mathrm{gate}_{t,s}(r_t^{\mathrm{CLS}}) =$
\begin{multline}
    \mathrm{softmax}(W_s^\mathrm{gate} \cdot r_t^{\mathrm{CLS}} + b_s^\mathrm{gate}) \in \mathbb{R}^5,
    \label{eq:gate}
\end{multline}
i.e., each slot gate is realized by a trainable linear layer classification head for BERT.

Boolean slots, i.e., slots that only take binary values, are treated separately. Here, the list of possible classes is $C_{\mathrm{bool}} = \{none, dontcare, true, false\}$ and the slot gate probability is $p^\mathrm{bgate}_{t,s}(r_t^{\mathrm{CLS}}) =$
\begin{multline}
    \mathrm{softmax}(W_s^\mathrm{bgate} \cdot r_t^{\mathrm{CLS}} + b_s^\mathrm{bgate}) \in \mathbb{R}^4.
    \label{eq:bgate}
\end{multline}

\subsection{Span-based Value Prediction}

For each slot $s$ that is to be filled via span prediction, a domain-slot specific span prediction layer takes the token representations $[r_t^1, \dots, r_t^{\mathrm{seq_{max}}}]$ of the entire dialog context for turn $t$ as input and projects them as follows:
\begin{subequations}
    \begin{align}
        [\alpha^s_{t,i}, \beta^s_{t,i}] &= W^\mathrm{span}_s \cdot r_t^i + b^\mathrm{span}_s \in \mathbb{R}^2 \\
        p^{\mathrm{start}}_{t,s} &= \mathrm{softmax}(\alpha^s_t) \\
        p^{\mathrm{end}}_{t,s} &= \mathrm{softmax}(\beta^s_t) \\
        \mathrm{start}^s_t &= \mathrm{argmax}(p^{\mathrm{start}}_{t,s}) \\
        \mathrm{end}^s_t &= \mathrm{argmax}(p^{\mathrm{end}}_{t,s}).
    \end{align}
\end{subequations}
Each span predictor is realized by a trainable linear layer classification head for BERT, followed by two parallel softmax layers to predict start and end position. Note that there is no special handling for erroneously predicting $\mathrm{end}^s_t < \mathrm{start}^s_t$. In practice, the resulting span will simply be empty.

\subsection{System Inform Memory for Value Prediction}

The system inform memory $I_t = \{I_t^1, \dots, I_t^N\}$ keeps track of all slot values that were informed by the system in dialog turn $t$. A slot in $DS_t$ needs to be filled by an informed value, if the user positively refers to it, but does not express the value such that span prediction can be used. E.g., in Figure~\ref{fig:diag}
the slot gate for domain-slot \texttt{<restaurant,name>} should predict $\mathit{inform}$. The slot is filled by copying the informed value into the dialog state, i.e., $DS_t^s = I_t^s$, where $s$ is the index of the respective domain-slot.

\subsection{DS Memory for Coreference Resolution}

The more complex a dialog can be, the more likely it is that coreferences need to be resolved. For instance, the name of a restaurant might very well be the destination of a taxi ride, but the restaurant might not be referred to explicitly upon ordering a taxi within the same conversation. Coreference resolution is challenging due to the rich variety of how to form referrals, as well as due to the fact that coreferences often span multiple turns. An example of a coreference that can be handled by our model is found in the example in Figure~\ref{fig:diag}.

The third copy mechanism utilizes the DS as a memory to resolve coreferences. If a slot gate predicts that the user refers to a value that has already been assigned to a different slot during the conversation, then the probability distribution over all possible slots that can be referenced is
\begin{multline}
    p^\mathrm{refer}_{t,s}(r_t^{\mathrm{CLS}}) =\\
    \mathrm{softmax}(W^s_\mathrm{refer} \cdot r_t^{\mathrm{CLS}} + b^s_\mathrm{refer}) \in \mathbb{R}^{N+1},
    \label{eq:refer}
\end{multline}
i.e., for each slot, a linear layer classification head either predicts the slot which contains the referenced value, or \emph{none} for no reference.

\subsection{Auxiliary Features}

Some recent approaches to neural DST utilize auxiliary input to preserve contextual information. For instance, SOM-DST adds the dialog state to its single-turn input as a means to preserve context across turns.

We already include contextual information in the input to BERT by appending the dialog history $H_t$. In addition to that, we also create auxiliary features based on the system inform memory and the DS memory. We generate two binary vectors $a_t^{\mathrm{inform}} \in \{0,1\}^N$ and $a_t^{\mathrm{ds}} \in \{0,1\}^N$ that indicate whether (1) a slot has recently been informed (based on the system inform memory), or (2) a slot has already been filled during the course of the dialog (based on the DS memory). These vectors are added to the output of BERT in a late fusion approach, and the slot gate probabilities in Equations~\ref{eq:gate},~\ref{eq:bgate} and~\ref{eq:refer} become $p^\mathrm{gate}_{t,s}(\hat{r}_t^{\mathrm{CLS}}),
p^\mathrm{bgate}_{t,s}(\hat{r}_t^{\mathrm{CLS}})$ and
$p^\mathrm{refer}_{t,s}(\hat{r}_t^{\mathrm{CLS}}),$
with $\hat{r}_t^{\mathrm{CLS}} = r_t^{\mathrm{CLS}} \oplus a_t^{\mathrm{inform}} \oplus a_t^{\mathrm{ds}}$.

\subsection{Partial Masking}

We partially mask the dialog history $H_t$ by replacing values with BERT's generic [UNK] token. The masking is partial in the sense that it is applied only to the past system utterances. For the system utterances, the contained values are known and their masking is straightforward. The idea behind partially masking the history is that the model is compelled to focus on the historical context information rather than the sighting of specific values. This should result in more robust representations $r_t^{\mathrm{CLS}}$ and therefore better overall slot gate performance.

\subsection{Dialog State Update}

We employ the same rule-based update mechanism as~\citet{chao2019bert} to track the dialog state across turns. At every turn, we update a slot, if a value has been detected which is not \emph{none}. If a slot-value is predicted as \emph{none}, then the slot will not be updated.

\section{Experimental Setup}

\subsection{Datasets}

We train and test our model on four datasets, MultiWOZ 2.1~\cite{eric2019multiwoz}, WOZ 2.0~\cite{wen2016network}, sim-M and sim-R~\cite{shah2018building}.
Among these, MultiWOZ 2.1 is by far the most challenging dataset. It is comprised of over 10000 multi-domain dialogs defined over a fairly large ontology. There are 5 domains (train, restaurant, hotel, taxi, attraction) with 30 domain-slot pairs that appear in all portions of the data.

The other datasets are single-domain and significantly smaller. Evaluations on these mainly serve as sanity check to show that we don't overfit to a particular problem. Some slots in sim-M and sim-R show a high out-of-vocabulary rate, making them particularly interesting for evaluating value independent DST.

The single domain datasets come with span labels. However, MultiWOZ 2.1 does not. We therefore generate our own span labels by matching the ground truth value labels to their respective utterances.

\subsection{Evaluation}

We compute the joint goal accuracy (JGA) on all test sets for straightforward comparison with other approaches. The joint goal accuracy defined over a dataset is the ratio of dialog turns in that dataset for which all slots have been filled with the correct value according to the ground truth. Note that \emph{none} needs to be predicted if a slot value is not present in a turn.
In addition to JGA, we compute the accuracy of the slot gates (joint and per-class) and various other metrics
for a more detailed analysis of model design decisions.

We run each test three times with different seeds and report the average numbers for more reliable results. 
MultiWOZ 2.1 is in parts labeled inconsistently. For a fair evaluation, we consider a value prediction correct, if it matches any of its valid labels (for instance "centre" and "center" for the slot-value \emph{hotel-area=centre}) as being correct. We semi-automatically analyzed value label inconsistencies in the training portion of the dataset in order to identify all label variants for any given value. During testing, these mappings are applied as is.

\begin{table}
    \centering
    \begin{tabular}{lr}
        \hline
        \textbf{Models} & \textbf{MultiWOZ 2.1} \\
        \hline
        DST-reader~\shortcite{gao2019dialog} & 36.40\% \\
        DST-span~\shortcite{zhang2019find} & 40.39\% \\
        SUMBT~\shortcite{lee2019sumbt} & 42.40\%$^{**}$ \\
        TRADE~\shortcite{wu2019transferable} & 45.60\% \\
        MA-DST~\shortcite{kumar2020ma} & 51.04\% \\
        DS-DST~\shortcite{zhang2019find} & 51.21\% \\
        SOM-DST~\shortcite{kim2019efficient} & 52.57\% \\
        DST-picklist~\shortcite{zhang2019find} & 53.30\% \\
        \hline
        TripPy & \textbf{55.29$\pm$0.28\%} \\
        \hline
    \end{tabular}
\caption{\label{tab:baselines_multiwoz}
DST Results on MultiWOZ 2.1 in JGA ($\pm$ denotes the standard deviation. $^{**}$ MultiWOZ 2.0 result.}
\end{table}

\subsection{Training}

We use the pre-trained \emph{BERT-base-uncased} transformer~\cite{vaswani2017attention} as context encoder front-end. This model has 12 hidden layers with 768 units and 12 self-attention heads each. The maximum input sequence length is set to 180 tokens after WordPiece tokenization~\cite{wu2016google}, except for MultiWOZ 2.1, where we set this parameter to 512. We compute the joint loss as
\begin{equation}
    \mathcal{L} = 0.8 \cdot \mathcal{L}_{\mathrm{gate}} + 0.1 \cdot \mathcal{L}_{\mathrm{span}} + 0.1 \cdot \mathcal{L}_{\mathrm{refer}}.
\end{equation}
The function for all losses is joint cross entropy. As there is no coreferencing in the evaluated single-domain datasets, the refer loss is not computed in those cases and the loss function is
\begin{equation}
    \mathcal{L} = 0.8 \cdot \mathcal{L}_{\mathrm{gate}} + 0.2 \cdot \mathcal{L}_{\mathrm{span}}
\end{equation}
instead.
Span predictors are presented only spans from the user utterances $U_i$ to learn from (including the user utterances in the history portion $H_i$ of the input). During training we set the span prediction loss to zero for all slots that are not labeled as \emph{span}. Likewise, the coreference prediction losses are set to zero if slots are not labeled as ~\emph{refer}. For optimization we use Adam optimizer~\cite{kingma2014adam} and backpropagate through the entire network including BERT, which constitutes a fine-tuning of the latter. The initial learning rate is set to $2e^{-5}$. We conduct training with a warmup proportion of 10\% and let the LR decay linearly after the warmup phase. Early stopping is employed based on the JGA of the development set.
During training we use dropout~\cite{srivastava2014dropout} on the BERT output with a rate of 30\%. We do not use slot value dropout~\cite{xu2014targeted} except for one dataset (sim-M), where performance was greatly affected by this measure (see Section~\ref{sec:results:ssec:analysis}).

\begin{table}
    \centering
    \begin{tabular}{lr}
        \hline
        \textbf{Models} & \textbf{WOZ 2.0} \\
        \hline
        NBT~\shortcite{mrkvsic2016neural} & 84.2\% \\
        BERT-DST~\shortcite{chao2019bert} & 87.7\% \\
        GLAD~\shortcite{zhong2018global} & 88.1\% \\
        GCE~\shortcite{nouri2018toward} & 88.5\% \\
        StateNet~\shortcite{ren2018towards} & 88.9\% \\
        SUMBT~\shortcite{lee2019sumbt} & 91.0\% \\
        \hline
        TripPy & \textbf{92.7$\pm$0.2\%} \\
        \hline
    \end{tabular}
\caption{\label{tab:baselines_dstc}
DST Results on WOZ 2.0.}
\end{table}

\begin{table}
    \centering
    \begin{tabular}{lrr}
        \hline
        \textbf{Models} & \textbf{sim-M} & \textbf{sim-R}\\
        \hline
        SMD-DST~\shortcite{rastogi2017scalable} & 96.8\%$^*$ & 94.4\%$^*$ \\
        \hline
        LU-DST~\shortcite{rastogi2018multi} & 50.4\% & 87.1\% \\
        BERT-DST~\shortcite{chao2019bert} & 80.1\% & 89.6\% \\
        \hline
        TripPy &  \textbf{83.5$\pm$1.2\%} & \textbf{90.0$\pm$0.2\%} \\
        \hline
    \end{tabular}
\caption{\label{tab:baselines_sim}
DST Results on sim-M and sim-R. $^*$ should be considered as oracle because the value candidates are ground
truth labels.}
\end{table}

\section{Experimental Results}

Tables~\ref{tab:baselines_multiwoz},~\ref{tab:baselines_dstc} and~\ref{tab:baselines_sim} show the performance of our model in comparison to various baselines. TripPy achieves state-of-the-art performance on all four evaluated datasets, with varying distance to the runner-up. Most notably, we were able to push the performance on MultiWOZ 2.1, the most complex task, by another 2.0\% absolute compared to the previous top scoring method, achieving 55.3\% JGA. The improvements on the much smaller datasets WOZ 2.0, sim-M and sim-R demonstrate that the model benefits from its design on single-domain tasks as well. The following analysis serves a better understanding of our model's strengths.

\subsection{Analysis}
\label{sec:results:ssec:analysis}

We analyse the performance of TripPy on ablation experiments on MultiWOZ 2.1 (see Table~\ref{tab:ablation}). 
Our baseline model is best compared to BERT-DST~\cite{chao2019bert}; we only take single turns as input, and only use span prediction to extract values from the turn. The resulting performance is comparable to other span-based methods such as DST-reader and DST-span and confirms that the dialogs in MultiWOZ are too complex to only be handled by this information extracting mechanism alone.

\begin{table}
    \centering
    \begin{tabular}{lr}
        \hline
        \textbf{Model} & \textbf{JGA} \\
        \hline
        Span prediction only (entire turn) & 42.63\%  \\
        \hline
        + triple copy mechanism & 49.23\% \\
        \quad + dialog history & 52.58\% \\
        \quad\quad + auxiliary features & 54.08\% \\
        \quad\quad\quad + masking & 54.29\% \\
        \hline
        TripPy (full sequence width) & 55.29\% \\
        \hline
    \end{tabular}
\caption{\label{tab:ablation}
Ablation experiments for our model.}
\end{table}

\begin{figure}
	\centering
	\includegraphics[page=1, trim=9cm 6.5cm 9cm 7.3cm, clip=true, width=1.00\linewidth,]{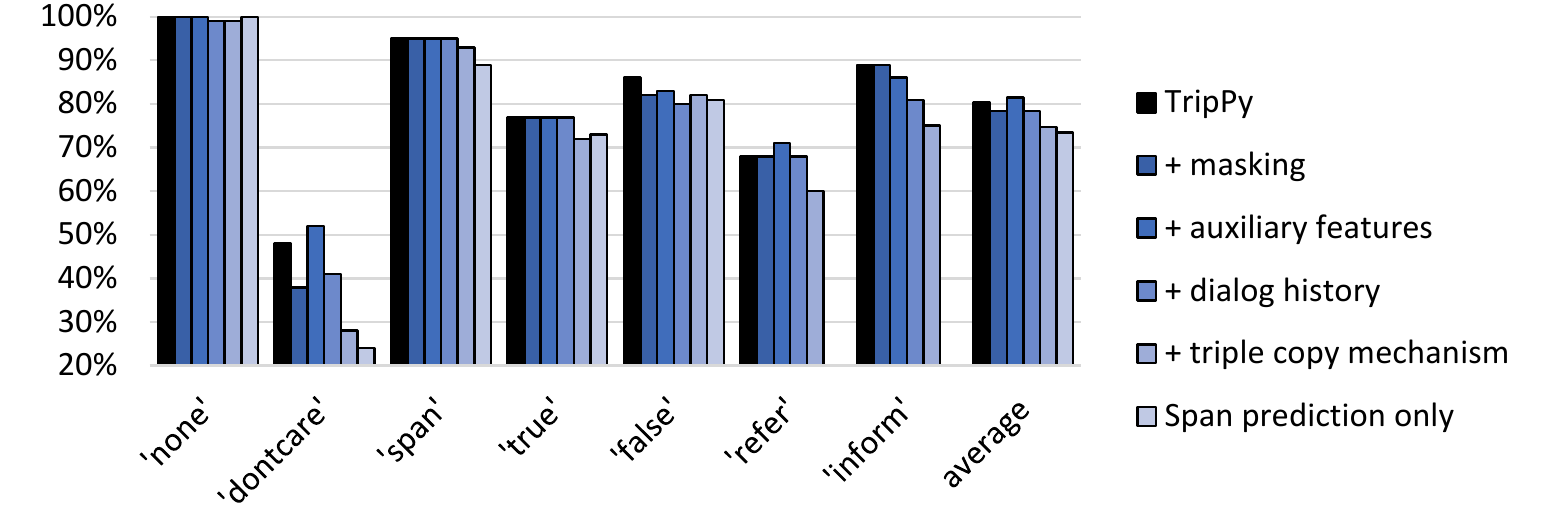}
	\caption{Per class performance of the slot gates for different versions of our model (ablation study).}
	\label{fig:slotgates}
\end{figure}

\paragraph{Impact of the triple copy mechanism}

Using our proposed triple copy mechanism pushes the performance close to 50\%, surpassing TRADE and closing in on the leading hybrid approaches. Especially the performance of the slot gates benefits from this change (see Figure~\ref{fig:slotgates}).
When looking at the F1 score for the individual classes, one can see that the \emph{span} class benefits from the distinction.
It is important to point out that none of the coreferences that our model handles can be resolved by span-prediction alone. This means that otherwise guaranteed misses can now be avoided and coreferences can be resolved by copying values between slots. What's more, using the dialog state memory to resolve coreferences helps value detection across multiple turns, as a value that has been referred to in the current turn might have been assigned to another slot multiple turns before.

\paragraph{Impact of the dialog history}

We found that using the dialog history as additional context information is critical to a good performance, as it reduces contextual ambiguity. This is clearly reflected in the improved performance of the slot gates (see Figure~\ref{fig:slotgates}), which has two positive effects. First, the presence and type of values is recognized correctly more often. Especially the special value \emph{dontcare}, and boolean slots (taking values \emph{true} and \emph{false}) benefit from the additional context. This is only logical, since they are predicted by the slot gate using the representation vector of the [CLS] token. Second, values are assigned to the correct slot more often than without the additional contextual information. With the additional dialog history, we outperform DS-DST and match SOM-DST, which set the previous state-of-the-art.

\begin{figure}[t]
	\centering
	\includegraphics[page=4, trim=9cm 3cm 9cm 4.1cm, clip=true, width=1.00\linewidth,]{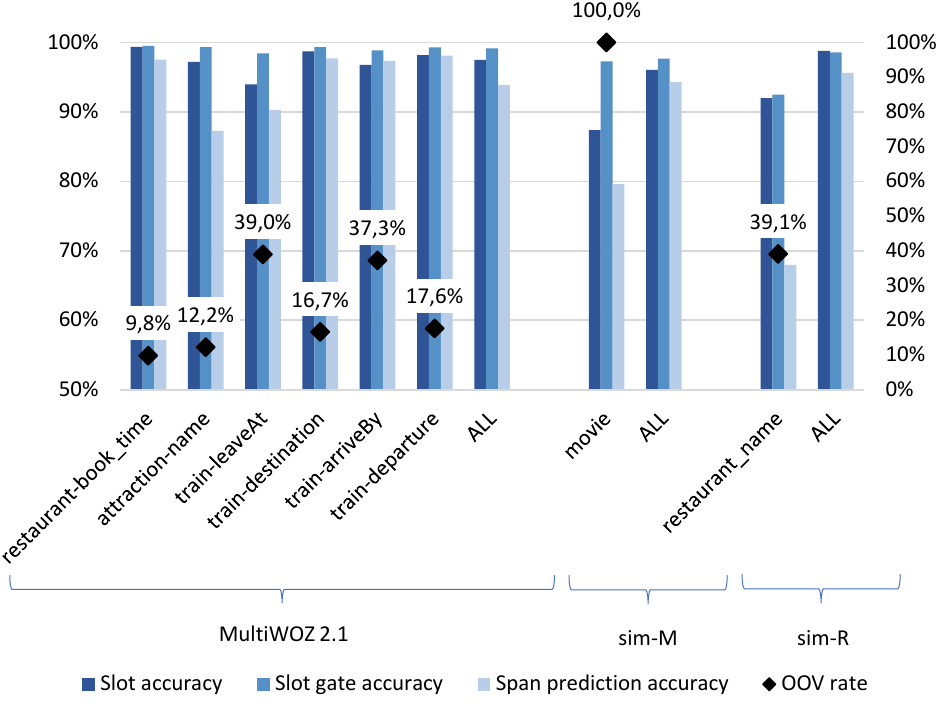}
	\caption{Performance of TripPy on slots with high OOV rate. \emph{ALL} denotes the average of all slots of the respective dataset.}
	\label{fig:oov_slots}
\end{figure}

\paragraph{Impact of the auxiliary features}

SOM-DST uses single turns as input, but preserves additional contextual information throughout the dialog by using the dialog state as auxiliary input. By adding our memory based auxiliary features in a late fusion approach, we surpass SOM-DST, and ultimately DST-picklist, which performs slot-filling with the knowledge of the full ontology.
Even though our features carry less information, that is, only the identities of the informed slots -- tracked by the system inform memory -- and the identities of the previously seen slots -- tracked by the DS memory --, we see substantial improvement using them. Obviously, more information about the progress of the dialog helps the slot gates and the referral gates in their classification tasks.

\paragraph{Impact of partial masking}

We found that masking the informed values in past system utterances does not give a clear benefit, but it also does not harm performance of the slot gates. While the \emph{inform} cases are detected more accurately, some other cases suffer from the loss of information in the input. Overall, there is a minor overall improvement observable. We report the numbers for MultiWOZ in Table~\ref{tab:ablation} and Figure~\ref{fig:slotgates}, but would like to note that we have seen the same trend on all other datasets as well.

\paragraph{Impact of the context width}

Our best model utilizes the full width of BERT (512 tokens). This is a clear advantage for longer dialogs. Maximal context width is not a decisive factor for the single-domain datasets, since their dialogs tend to be shorter. As expected, we have not seen any change in performance on these. For MultiWOZ, we gain 1\% absolute by maximizing the history length to preserve as much of the dialog history as possible, achieving 55.3\% JGA.

\begin{figure}[t]
	\centering
	\includegraphics[page=2, trim=11.8cm 6.5cm 11.7cm 7.2cm, clip=true, width=1.00\linewidth,]{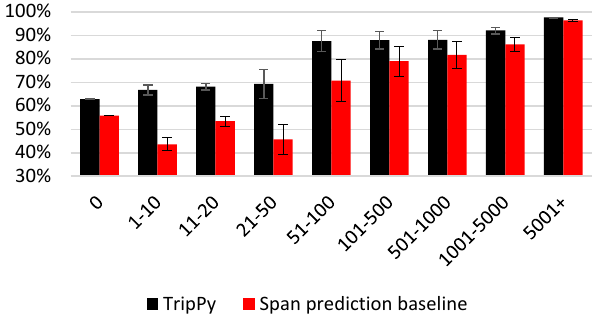}
	\caption{Recall of values depending on the amount of samples seen during training. 0 seen samples means the value is OOV during test time.}
	\label{fig:oovs}
\end{figure}

\subsection{Generalization Study}
\label{sec:results:ssec:generalization}

It is important that a DST model generalizes well to previously unseen values. We looked at the performance of our model on slots with exceptionally high out-of-vocabulary rates, of which we identified 8 across the evaluated datasets. Figure~\ref{fig:oov_slots} plots performance measures for these slots and compares them to the average performance for all slots in the respective datasets. Generally, the slots that expect named entities as values show the lowest accuracy. However, the below-average performance of these slots does not seem to be caused by a particularly high OOV rate. Even at 100\%, the \emph{movie} slot of sim-M still performs comparably well. Other slots with relatively high OOV rate still perform close to or better than the average.

\begin{figure}[t]
	\centering
	\includegraphics[page=3, trim=9.8cm 4cm 10cm 4.5cm, clip=true, width=1.00\linewidth,]{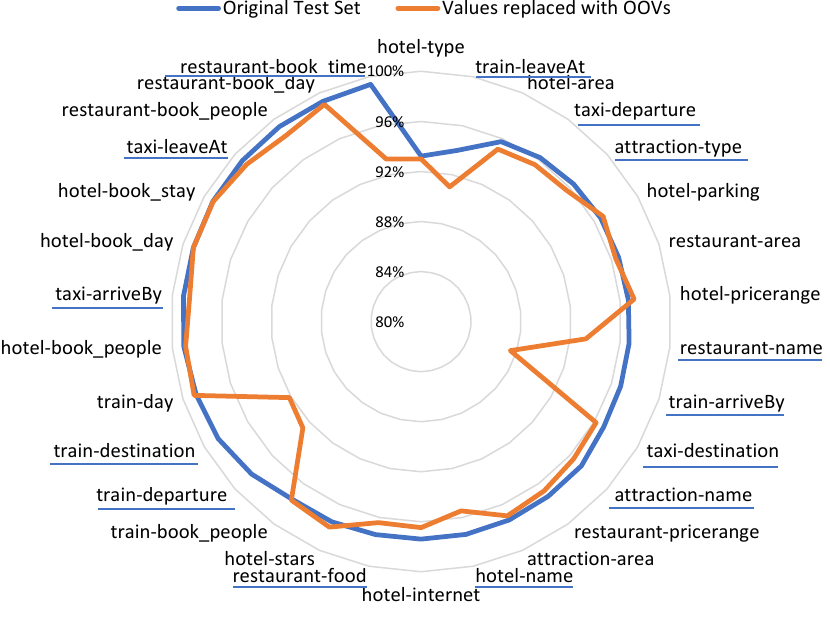}
	\caption{Per-slot accuracy of TripPy on the original test set and the OOV test set. Underlined slot names indicate slots with at least one OOV value.}
	\label{fig:oov_circle}
\end{figure}

Figure~\ref{fig:oovs} plots the recall of values depending on the number of samples seen during training. To our surprise, it does not seem to matter whether a particular value has never been seen during training in order to be detected correctly. OOV values are detected just as well as generally less common values. Our observations however indicate that the model benefits tremendously by seeing a certain minimal amount of training samples for each value, which is somewhere around 50. In other words, if such amounts of data are available, then the model is able to effectively utilize them. In the same Figure we compare TripPy to the span prediction baseline. The latter clearly struggles with OOVs and rare values and generally seems to require more training samples to achieve a good recall. The higher recall on OOV values is likely caused by the fact that many unseen values are of the category time of day, which mostly follows a strict format and is therefore easier to spot. Overall, TripPy clearly generalizes better over sample counts.

To test the limits of our model's generalization capacities, we manually replaced most of the values in the MultiWOZ test set by (fictional but still meaningful) OOV values. Of the over 1000 unique slot-value pairs appearing in the modified test set, about 84\% are OOV after the replacement. Figure~\ref{fig:oov_circle} compares the per-slot accuracy of our model on the original test set and the OOV test set. Underlined slot names indicate slots with at least one OOV value. Their average OOV rate is 90\%. Surprisingly, most of these slots maintain their high accuracy and only few suffer from the high OOV count. Mainly it is one particular domain, \emph{train}, which suffers above-average performance drops. However, the remainder of the slots maintain their performance. This demonstrates that our model is well equipped to handle OOV values, regardless of the type (e.g., named entity, time of day).

\section{Conclusion}

We have demonstrated that our approach can handle challenging DST scenarios. Having to detect unseen values does not considerably impair our model's general performance. The information extraction capabilities of our proposed model are rooted in the memory-based copy mechanisms and perform well even in extreme cases as discussed in Section~\ref{sec:results:ssec:generalization}. The copy mechanisms are not limited by a predefined vocabulary, since the memories themselves are value agnostic.

To further improve the DST capabilities of TripPy, we hope to introduce slot independence as at present its tracking abilities are limited to slots that are predefined in the ontology. For that, We would like to expand our approach towards the schema-guided paradigm for dialog modeling. We also would like to employ a more sophisticated update strategy, for example by adding the option to partially forget. There already exists an intriguing set of works focusing on these issues and we hope to incorporate and expand upon it in the future.

\section*{Acknowledgments}

M. Heck, C. van Niekerk and N. Lubis are supported by funding provided by the Alexander von Humboldt Foundation in the framework of the Sofja Kovalevskaja Award endowed by the Federal Ministry of Education and Research, while C. Geishauser, H-C. Lin and M. Moresi are supported by funds from the European Research Council (ERC) provided under the Horizon 2020 research and innovation programme (Grant agreement No. STG2018\_804636).

\bibliography{sigdial2020}
\bibliographystyle{acl_natbib}

\end{document}